# A Physics-Guided AI Cascaded Corrector Model Significantly Extends Madden-Julian Oscillation Prediction Skill

Xiao Zhou[a†], Yuze Sun[a†], Jie Wu[b], Xiaomeng Huang[a]

a Department of Earth System Science, Ministry of Education Key Laboratory for Earth System Modelling, Institute for Global Change Studies, Tsinghua University, Beijing, China

b State Key Laboratory of Climate System Prediction and Risk Management/China Meteorological Administration Key Laboratory for Climate Prediction Studies, National Climate Centre, China Meteorological Administration, Beijing, 100081

† These authors contributed equally to this work.

# Abstract

The Madden-Julian Oscillation (MJO) is an important driver of global weather and climate extremes, but its prediction in operational dynamical models remains challenging, with skillful forecasts typically limited to 3-4 weeks. Here, we introduce a novel deep learning framework, the Physics-guided Cascaded Corrector for MJO (PCC-MJO), which acts as a universal post-processor to correct MJO forecasts from dynamical models. This two-stage model first employs a physics-informed 3D U-Net to correct spatial-temporal field errors, then refines the MJO's RMM index using an LSTM optimized for forecast skill. When applied to three different operational forecasts from CMA, ECMWF and NCEP, our unified framework consistently extends the skillful forecast range (bivariate correlation > 0.5) by 2–8 days. Crucially, the model effectively mitigates the "Maritime Continent barrier," enabling more realistic eastward propagation and amplitude. Explainable AI analysis quantitatively confirms that the model's decision-making is spatially congruent with observed MJO dynamics (correlation > 0.93), demonstrating that it learns physically meaningful features rather than statistical fittings. Our work provides a promising physically consistent, computationally efficient, and highly generalizable pathway to break through longstanding barriers in subseasonal forecasting.

# Plain Language Summary

Predicting the weather beyond two weeks is a major challenge for forecasters. A key player in this timescale is the Madden-Julian Oscillation (MJO), a large-scale climate pattern that moves slowly across the tropics and influences extreme weather worldwide, from hurricanes to heatwaves. However, the best computer models struggle to predict the MJO accurately for 3-4 weeks. We developed a new artificial intelligence (AI) tool that acts like a powerful "spell-checker" for these dynamical models. It takes their MJO forecasts and corrects the errors. When we tested it on three widely used operational forecasting systems, this AI corrector added 2-8 days of additional skillful forecast time. A major success was that it helped the MJO propagation over the Maritime Continent—a group of islands in Indonesia that often blocks the MJO in forecasts, causing it to stall and die. Importantly, we used special techniques to peer inside the AI's "black box" and confirmed that it truly learned the real physics of the MJO, making its corrections trustworthy. This AI tool provides a fast and effective way to make our long-range forecasts more reliable.

## Key Points

A physics-guided design post-processor extends the skillful forecast limit of the Madden-Julian Oscillation by 2–8 days across three operational models.

The model's physics-guided design and explainable AI analysis confirm it learns authentic MJO dynamics, not just statistical fittings.

The framework effectively mitigates the "Maritime Continent barrier," enabling more realistic eastward propagation in forecasts.

# 1 Introduction

The Madden-Julian Oscillation (MJO), first identified by Madden and Julian (1971, 1972), is widely regarded as the primary mode of intraseasonal variability in the tropics and a key source of subseasonal predictability (Zhang, 2005, 2013). It features planetary-scale, eastward-moving patterns of enhanced and suppressed deep convection and circulation, with a typical period of 30–90 days, especially noticeable during boreal winter (Adames & Wallace, 2014; Lau & Waliser, 2012; Li et al., 2020). Through direct local effects and the excitation of Rossby wave trains, the MJO greatly influences global weather and climate extremes, including monsoons, tropical cyclones, heatwaves, and atmospheric rivers (Cassou, 2008; Henderson et al., 2017; Zhang et al., 2018; Stan et al., 2022). Therefore, accurately predicting the MJO is a crucial benchmark for subseasonal-to-seasonal (S2S) forecasting, as better MJO predictions have been shown to extend the forecast range for global weather and climate anomalies (Lin et al., 2009; Vitart & Robertson, 2018; Guo et al., 2025).

Operational dynamical models remain the primary tool for MJO prediction, yet their skill is often limited to 3–4 weeks (Jiang et al., 2020; Vitart et al., 2025), plagued by issues such as the "Maritime Continent barrier" and rapid amplitude decay (Kim et al., 2018; Wang et al., 2018). To extend the forecast skill, statistical post-processing techniques—such as the Linear Inverse Model (LIM; Penland & Magorian, 1993), Principal Oscillation Pattern (POP; Hasselmann, 1988) analysis, and multiple linear regression—are commonly employed to correct model outputs. For example, Wu and Jin (2021) applied a LIM-based approach to rectify biases in the MJO's linear dynamics, achieving a skill extension of approximately 2–4 days. However, this paradigm struggles to fully capture and correct the complex nonlinear physical processes and large-scale spatial-temporal coherence inherent in MJO evolution, thus limiting its potential for further improving forecast skill.

Recent advances in artificial intelligence (AI) have opened new avenues for MJO forecasting, which can be broadly categorized into two paradigms. The first is a pure AI approach, which forecasts the MJO directly from initial atmospheric states or historical observations, bypassing dynamical models entirely (e.g., Chen et al., 2024). While some studies report competitive skill, this paradigm often requires substantial computational resources for training and inference, posing a barrier to operational deployment. Moreover, it fails to leverage the physical priors and forecast information embedded in state-of-the-art dynamical models, and often leads to over-smoothed outcomes at longer lead times. The second paradigm employs AI as a post-processor to correct biases in the low-dimensional RMM indices derived from dynamical forecasts (Kim et al., 2021; Silini et al., 2022). This approach effectively reduces average amplitude and phase errors across multiple models and can even elevate the skill of poorer models to match that of the best ones. However, this "index-to-index" correction operates in a highly reduced space, incapable of providing physically consistent, spatially explicit field forecasts (e.g., OLR, zonal wind). Furthermore, its decision-making process often lacks physical interpretability, and its performance may depend on the availability of multi-model ensembles.

To address these limitations, we introduce a third paradigm: a physics-guided, cascaded deep learning framework that serves as a universal post-processor for MJO forecasting. This approach is designed to correct the full spatial-temporal fields of key MJO-related variables (e.g., OLR, U850, U200) from operational dynamical models, establishing a physically consistent basis for improved prediction. The corrected fields are subsequently projected onto the MJO phase space to refine the RMM indices. This "field-to-field-to-index" architecture ensures that enhancements in the low-dimensional RMM index are rooted in coherent, high-dimensional physical processes. By leveraging existing dynamical model forecasts and incorporating physical constraints, our framework aims to significantly extend the skillful forecast range while providing trustworthy and interpretable corrections, thereby overcoming the key limitations of previous AI-based approaches.

The remainder of this paper is structured as follows. Section 2 describes the data and methodology. Section 3 evaluates the forecast skill improvements and analyzes the model architecture. We also provide a physical interpretation of the model in this section. Finally, Section 4 summarizes the conclusions and outlines potential directions for future work.

# 2 Data and Method

## 2.1 Data

This study utilizes subseasonal forecast data from three widely used operational centers from S2S Project Phase II dataset （Vitart et al., 2017, 2018, 2025）: the Beijing Climate Center of China Meteorological Administration (CMA; Wu et al., 2021), the European Center for Medium-Range Weather Forecasts (ECMWF), and the National Centers for Environmental Prediction (NCEP; Zhu et al., 2018). Detailed configurations of these prediction systems are outlined in Table 1. Noted that some ECMWF data are unavailable in the achieving dataset, but the current forecast cases are sufficient to train stable AI post-processor model. This multi-model data is designed to rigorously evaluate the generalization capability of the physics-guided design correction framework in addressing systematic errors across different dynamical forecasting systems. The observational circulation and outgoing longwave radiation (OLR) reference is obtained from the NCEP/DOE Reanalysis II (NCEPII; Kanamitsu et al., 2002) and National Oceanic and Atmospheric Administration (NOAA, Liebmann & Smith, 1996). For both model forecasts and reanalysis, we use daily fields of

OLR and zonal wind at 850 hPa (U850) and 200 hPa (U200), focusing on the equatorial region (20°S–20°N) and a forecast lead time of 40 days. All data are remapped to a uniform 2.5°×2.5° latitude-longitude grid (144×17 grid points).

Table 1. Configurations of S2S dynamical prediction systems.

| Data source | Forecast time (day) | Forecast frequency | Forecast period | Training Samples | Test Samples | Model version |
| --- | --- | --- | --- | --- | --- | --- |
| CMA | 60 | 2/week | 2006-2021 | 1331 | 333 | BCC-CSM2-HR |
| ECMWF | 46 | 2/week | 2001-2021 | 1092 | 273 | CY46R1 |
| NCEP | 44 | 1/day | 1999-2022 | 1604 | 401 | NCEP ensemble |

## 2.2 Preprocessing

To isolate intraseasonal variability associated with the MJO, we remove the seasonal cycle and interannual signals following established procedures (e.g., Lin et al., 2008; Wu & Jin, 2021). The daily climatology (1981–2010) is constructed from the first three harmonics and subtracted from the raw data. A 120-day running mean is then applied to filter out lower-frequency variability such as ENSO, yielding the final subseasonal-scale anomaly fields for model training and evaluation. Prior to model training, all input variables (OLR, U850, U200) from both dynamical forecasts and NCEPII reanalysis are standardized using the Z-score method to ensure consistent data distributions, which is essential for stable and efficient neural network training.

## 2.3 Verification Methodology

Forecast skill for the MJO is quantitatively assessed using the following metrics (Lin and Brunet, 2008) of the Real-time Multivariate MJO (RMM) index (Wheeler & Hendon, 2004). The correlation skill (COR), a standard metric for MJO forecast evaluation, measures the linear relationship between forecasted and observed RMM indices and is defined as:

$$COR(t) = \frac{\sum_{i=1}^{N} [a_{1i}(t) \cdot b_{1i}(t) + a_{2i}(t) \cdot b_{2i}(t)]}{\sqrt{\sum_{i=1}^{N} [a_{1i}^2(t) + a_{2i}^2(t)]} \cdot \sqrt{\sum_{i=1}^{N} [b_{1i}^2(t) + b_{2i}^2(t)]}}$$

where $a_{1i}(t)$ and $a_{2i}(t)$ are the observed RMM1 and RMM2 at day $t$, and $b_{1i}(t)$ and $b_{2i}(t)$ are their respective forecasts, for the $i$th forecast with a $t$-day lead. Here, $N$ is the number of forecasts. The root mean square error (RMSE) can be written as:

$$RMSE(t) = \sqrt{\frac{1}{N}\sum_{i=1}^{N}\{[a_{1i}(t) - b_{1i}(t)]^2 + [a_{2i}(t) - b_{2i}(t)]^2\}}$$

It takes into account errors in both phase and amplitude. The mean square skill score (MSSS) is defined as:

$$MSSS(t) = 1 - \frac{MSE_f(t)}{MSE_c}$$

where

$$MSE_f(t) = \frac{1}{N}\sum_{i=1}^{N}\{[a_{1i}(t) - b_{1i}(t)]^2 + [a_{2i}(t) - b_{2i}(t)]^2\}$$

is the mean square error of the model forecast and

$$MSE_c = \frac{1}{N}\sum_{i=1}^{N}[a_{1i}^2(t) + a_{2i}^2(t)]$$

is the climatological variance. MSSS provides a relative level of skill for the MJO forecast compared to a climatological forecast that predicts no MJO signal.

A forecast is considered skillful when the COR exceeds 0.5 (Wu & Jin, 2021). The effective forecast skill is defined as the lead time at which the COR first falls below this threshold. Statistical significance of skill differences between prediction methods is assessed using Steiger's Z-test (Raghunathan et al., 1996) at the 95% confidence level.

## 2.4 Model Design: A Physics-Guided Deep Learning Framework for MJO Correction

Accurate MJO prediction requires correcting errors in both the spatial structure and temporal evolution of forecast fields. To address this challenge, we propose the Physics-guided Cascaded Corrector for MJO (PCC-MJO), a novel architecture that operates in two dedicated stages. First, a spatial correction module establishes a physically consistent, large-scale background—a prerequisite for realistic MJO propagation. Subsequently, a temporal refinement module specifically targets the MJO signal itself by projecting the corrected fields onto the dominant MJO empirical orthogonal function (EOF) modes (Wheeler & Hendon, 2004) to obtain a preliminary RMM index, which is then further refined. This cascaded design forces the model to specialize, first in correcting general field errors and then in capturing MJO-specific dynamics.

The spatial correction is built upon a 3D U-Net (Ronneberger et al., 2015), chosen for its proven efficacy in tasks requiring precise, pixel-wise regression across multiple scales. Its encoder-decoder structure with skip connections is ideally suited to capture the multi-scale nature of the MJO, from planetary-scale circulations to organized convective clusters. We specifically tailor the network to the physics of the problem: the spatial convolutional kernels in the encoder (7,5), (5,3), (3,3), (3,3) are designed to hierarchically extract features from planetary to convective scales, a structure that is symmetrically reversed in the decoder. The temporal kernels (3, 7, 15, 21) are explicitly tuned to the sub-harmonics of the MJO life cycle. Ablation studies confirm that this targeted design is critical, as it significantly outperforms standard, scale-agnostic kernels (Supplementary Figures S1, S2).

The temporal refinement module introduces a physics-guided feature reduction by projecting the corrected fields onto the observed MJO EOF modes. This step acts as a dynamics-purifying bottleneck, effectively creating an "MJO filter" that focuses subsequent learning exclusively on the oscillation's core dynamics. The derived RMM index is processed by an LSTM network (Hochreiter & Schmidhuber, 1997) to capture the MJO's propagation characteristics. Crucially, we transition the optimization target from pixel-level reconstruction (Mean Squared Error) to mode-level forecast skill by directly maximizing the bivariate correlation coefficient. This strategic shift explicitly prioritizes the accuracy of the MJO's phase and amplitude—the key metrics in operational forecasting.

In summary, our model integrates a physics-aware U-Net and a skill-focused LSTM. This cascaded design effectively disentangles the learning of general background errors from that of MJO-specific biases, simplifying the overall regression task. Ablation experiments demonstrate a strong synergistic effect between the components; the removal of either leads to a substantial and statistically significant drop in forecast skill, quantitatively validating our architectural philosophy. The following results demonstrate how this targeted architecture translates into significant gains in forecast skill and physical consistency.

Complementing its skillful performance, our model is designed to be computationally efficient. The entire framework is lightweight, comprising 124,310,083 parameters in the 3D U-Net and only 34,923 parameters in the LSTM network. This efficiency enables rapid training, with the model converging within 2–4 hours on a single A100 GPU. More importantly, the inference for bias correction is exceptionally fast, generating a full set of corrected 40-day forecasts from raw model outputs in minutes on consumer-grade graphics hardware (e.g., an NVIDIA RTX 3090/4090). This stands in stark contrast to the immense computational cost of dynamical model integration, underscoring the practical potential of our physics-guided design post-processor for operational forecasting. The following results demonstrate how this targeted and efficient architecture translates into significant gains in forecast skill and physical consistency.

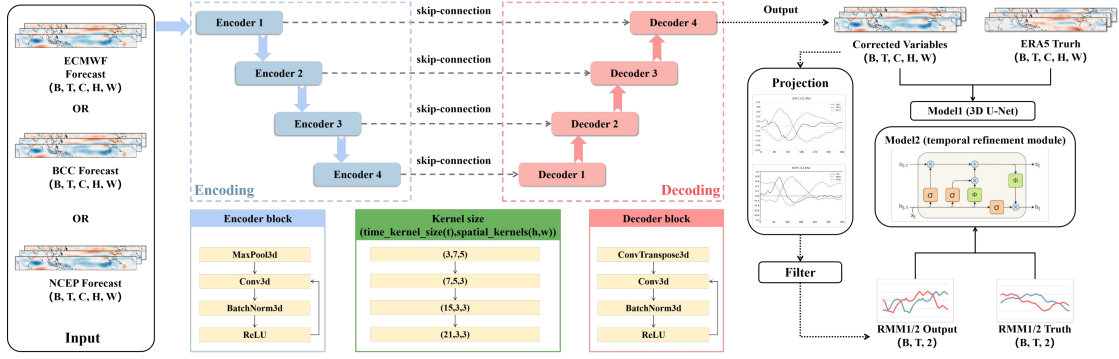

Figure 1. Schematic of the PCC-MJO architecture. The model consists of two main modules: 1) A 3D U-Net-based Spatial Reconstruction Module that takes the biased forecast fields (OLR, U850, U200) across lead times as input and outputs corrected physical fields. Its encoder (downsampling) and decoder (upsampling) paths are composed of 3D convolutional blocks with specifically designed spatial and temporal kernels to capture multi-scale MJO features. Skip connections preserve fine-grained details. 2) A Projection and LSTM-based Refinement Module where the reconstructed fields are projected onto the observed MJO EOF modes (Wheeler and Hendon, 2004) to obtain a preliminary RMM index. An LSTM network then further corrects this index sequence to produce the final, skillful RMM forecast.

The model was implemented using the PyTorch deep learning framework and trained on a single NVIDIA A100 Tensor Core GPU. We employed the Adam optimizer (Kingma & Ba, 2014) with a fixed learning rate of $1 \times 10^{-3}$ and a batch size of 32. The training process typically converged within 2–4 hours less than 100 epochs under this configuration. This efficient training setup, combined with the lightweight architecture described in Section 3.1, underscores the practical feasibility of our approach.

# 3 Results

## 3.1 Forecast Skill Improvement and Physical Consistency

A critical finding of this study is the demonstrated generalizability of our AI correction framework. As shown in Figure 2, the model delivers a substantial and statistically significant enhancement in MJO forecast skill across three independent dynamical models, each with distinct systematic errors and initial skill levels. The raw forecasts from CMA, ECMWF, and NCEP exhibit maximum skillful lead times (COR ≥ 0.5) of 33 days, 18 days, and 18 days, respectively. After processing through our unified model architecture, the skillful forecast range is consistently extended to 39 days, 26 days, and 20 days, respectively. This corresponds to a net improvement of 6, 8, and 2 days, with the most substantial gain for CMA and a statistically significant extension for NCEP. Critically, this robust improvement—achieved using a single correction model architecture—highlights its powerful capability to learn and correct diverse, model-specific biases. The framework effectively acts as a universal post-processor, adding significant value to both high- and low-skill baseline forecasts and underscoring its potential for operational application across different forecasting systems.

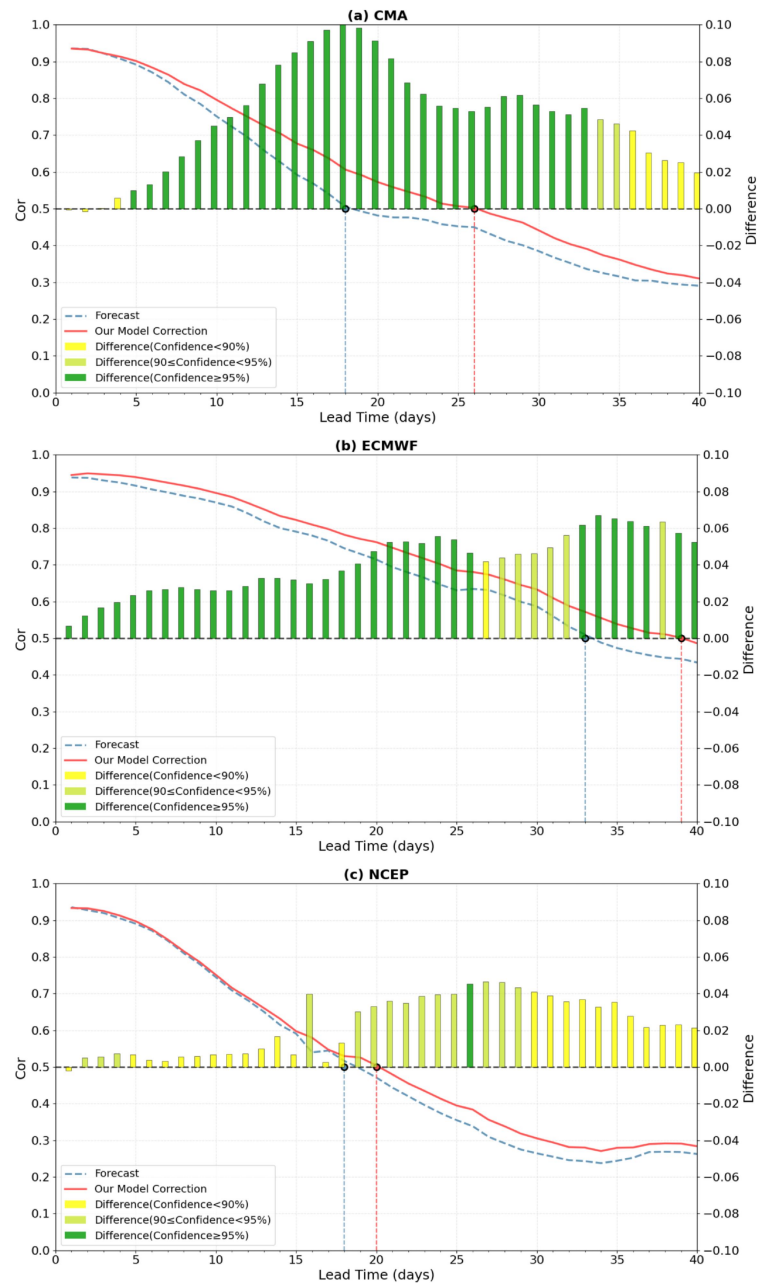

Figure 2. Enhancement of MJO forecast skill by the physics-guided design correction model. Bivariate anomaly correlation coefficient (COR) of the RMM index as a function of forecast lead time for (a) CMA, (b) ECMWF, and (c) NCEP forecasts. The blue dashed and red solid lines denote the skill of the raw dynamical forecasts and the AI-corrected forecasts, respectively. The vertical dashed line indicates the maximum skillful lead time (COR ≥ 0.5). Bar plots show the difference in COR between the corrected and raw forecasts, with shading indicating the statistical significance level of the improvement assessed by Steiger's Z-test (light hatching: 90–95%; dense hatching: ≥95%). All results are computed from the test dataset (the last 20% of the data).

The robust improvement shown in Figure 2 is further deconstructed for different MJO initial conditions and seasons (Supplementary Figures S3 and S4), which reveal under when and what conditions the AI correction is most impactful. Notably, the difference plots (shading) in both the season-dependent and phase-dependent evaluations exhibit their maximum improvements at lead times between 15 and 30 days. This period precisely coincides with the steep decline in COR of the raw forecasts (Figure 2a-c), where traditional models lose skill most rapidly. Our AI model provides the strongest corrective boost exactly at this critical juncture, thereby effectively flattening the skill-decay curve and directly explaining the extension of the skillful forecast range by 2-8 days. This synchronized enhancement across metrics and lead times underscores that the model intelligently targets the most vulnerable period in MJO forecasts, ensuring robust improvement regardless of the initial season or MJO phase.

Beyond the aggregate skill metrics, it is critical to evaluate whether the model re-produces the physically realistic evolutions of the MJO. We use ECMWF forecasts as a representative example to illustrate the correction in MJO propagation. Figure 3 composites the RMM phase-space trajectories for forecasts initialized with strong MJO events (amplitude > 1) across all eight initial phases. A key and consistent observation is that the AI-corrected forecasts (red trajectories) overwhelmingly lie closer to the observed evolution (black trajectories) than the raw forecasts (blue) across most phases. This demonstrates a systematic improvement in capturing the MJO's propagation pathway.

The correction is particularly crucial and physically meaningful when overcoming the "Maritime Continent (MC) barrier"—a longstanding challenge in MJO forecasting where models consistently fail to maintain the coherent eastward propagation of convective signals from the Indian Ocean to the Western Pacific (Hsu & Lee, 2005; Kim et al., 2018; Zhang et.al., 2017). This barrier is often attributed to models' difficulties in representing the intricate air-sea interactions and the triggering of deep convection over the complex topography of the Maritime Continent, leading to a premature dissipation or stagnation of the MJO in phases 4-5. As clearly seen in Figure 3, the raw ECMWF forecasts (blue) exhibit this classic bias, with trajectories frequently weakening and stalling over the MC region when initialized from Phase 3-4. In stark contrast, our AI-corrected forecasts (red) successfully sustain the MJO's eastward propagation, maintaining a more realistic amplitude and carrying the convective signal coherently into the Western Pacific. This indicates that our model has learned to correct not just the amplitude but, more importantly, the fundamental dynamical and physical process errors associated with the Maritime Continent barrier.

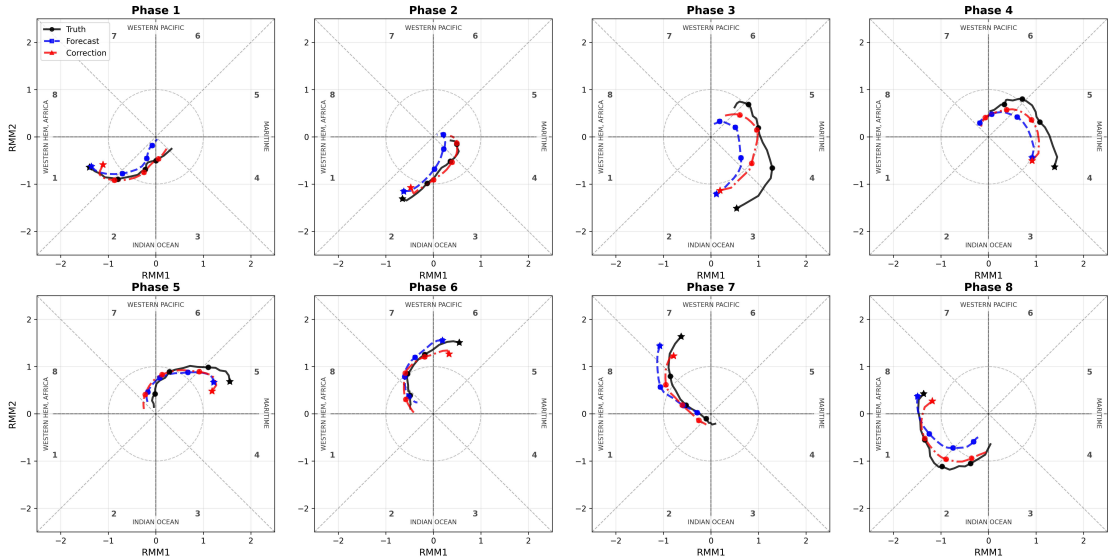

Figure 3. Improvement in MJO phase evolution forecasting. Composite RMM phase diagrams for forecasts initialized from strong MJO events (amplitude > 1) in different initial phases (1–8) for ECMWF forecast. The black, blue, and red trajectories represent the observed (NCEPII), raw forecast, and AI-corrected forecast evolutions, respectively. All trajectories start from the composite initial location for each phase. The AI correction consistently pulls the forecasted MJO evolution closer to the observed path, more accurately capturing both its propagation and amplitude.

The case studies in Supplementary Figures S5 and S6 further illustrate how our AI correction improves MJO forecasts in a physically consistent manner, instead of simply increasing amplitude. In Supplementary Figures S5, the raw forecast consistently underestimates MJO strength when passing through MC area, while in Supplementary Figures S6, it overestimates amplitude in Phases 3–4 but underestimates it later. Our model accurately rectifies these contrasting biases—strengthening where the forecast is too weak and reducing where it is too strong—producing a balanced evolution closer to observations. This nuanced correction is reflected in the consistently higher pattern correlation coefficients of the corrected OLR fields, confirming that the model not only adjusts amplitude but also better captures the realistic eastward propagation and spatial-temporal structure of the MJO.

## 3.2 Physical Interpretation via Explainable AI

To ensure the reliability and generalizability of our AI-based correction model, it is imperative to move beyond the "black box" paradigm and rigorously interpret its decision-making process. Only by demonstrating that the model has learned physically meaningful patterns—rather than merely performing a mathematical fit to the training data—can we build confidence in its ability to perform robustly on independent, unseen data. Explainable AI (XAI) analysis thus serves not only to validate the physical consistency of the corrections but also to establish the model's trustworthiness for future operational applications.

We achieve this through the Integrated Gradients method (Sundararajan et al., 2017), which quantifies the contribution of each input variable at every grid point to the final forecasted RMM indices. The method computes the gradient of the model's output (RMM1 and RMM2 separately) with respect to its input—the full 3D spatial-temporal fields of OLR, U850, and U200—along a path from a neutral baseline to the actual input values. The resulting attribution maps highlight regions where the model is most sensitive; larger absolute values indicate features that are crucial for its correction. We compute these maps for all test samples and composite them by the two leading MJO modes. The physical authenticity of the model is then quantitatively assessed by examining the spatial alignment between these high-sensitivity regions and the canonical structures of the observed MJO.

Figure 4 presents the composited attribution patterns for the forecasted (a) RMM1 and (b) RMM2 indices. The two-dimensional spatial maps (top three rows in each panel) reveal that the model's attention is not diffused but is strategically focused on key dynamical centers of action. Crucially, when these attribution maps are meridionally averaged over the tropical belt (15°S–15°N) to produce the line plots (bottom row), a remarkable alignment with the observed canonical MJO structures emerges (cf. Supplementary Figure S7). The longitudinal distribution of the model's attention for OLR, U850, and U200 exhibits a striking resemblance to the corresponding observed EOF patterns in both phase and amplitude. This visual correspondence is quantitatively confirmed by calculating the spatial correlation between the meridional-mean attribution pattern and the observed EOF pattern for each index. The correlation coefficients reach 0.94 for RMM1 and 0.93 for RMM2 (exceeding the 99% confidence level), unequivocally demonstrating that the model's decision logic is spatially congruent with the observed MJO dynamics. Furthermore, the model correctly captures the quadrature phase relationship between the RMM1 and RMM2 indices, as well as the established lead-lag relationships between the convective (OLR) and circulatory (U850, U200) variables. This compelling coherence demonstrates that the AI model has not learned a superficial statistical correction but has instead internalized the fundamental spatial dynamics of the MJO, prioritizing the same physical processes that define the oscillation in nature.

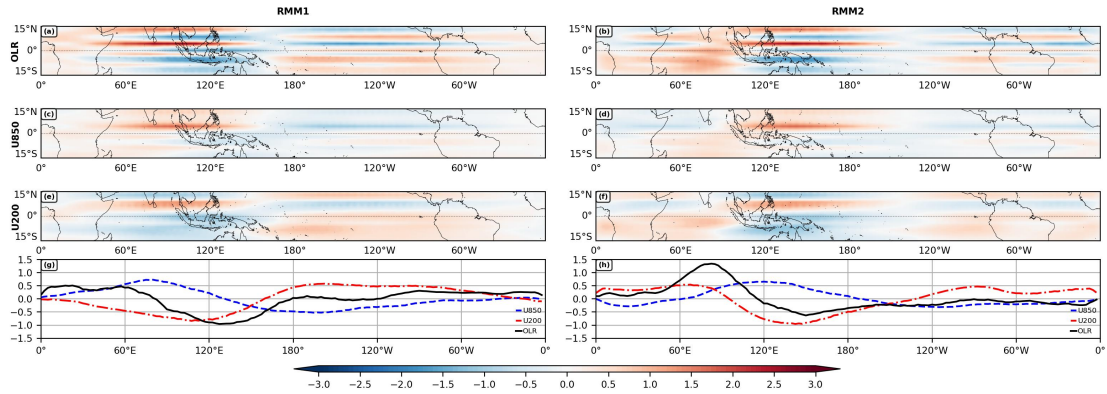

Figure 4. Physical interpretation of the AI model's decision-making using Integrated Gradients. Phase-composited attribution maps for the forecasted (a) RMM1 and (b) RMM2 indices. For each panel, the top three rows show the two-dimensional attribution patterns for outgoing longwave radiation (OLR), 850-hPa zonal wind (U850), and 200-hPa zonal wind (U200), respectively. The bottom row displays the meridional mean (15°S–15°N) of these attribution values, illustrating the longitudinal focus of the model's attention. The high spatial correlation between these attribution patterns and the observed MJO EOF modes (Supplementary Figure S7) quantitatively confirms the model's physical consistency.

## 4 Summary and Discussion

This study presents a novel deep learning framework that demonstrates substantial success in correcting and enhancing MJO forecasts from operational dynamical models. Our primary contribution is the development of a physically-guided, cascaded architecture that consistently extends the skillful forecast range (COR > 0.5), with notable improvements of 6, 8, and 2 days for CMA, ECMWF, and NCEP forecasts, respectively, demonstrating the framework's ability to add value even to more challenging forecasts. This level of improvement surpasses that typically achieved by traditional statistical correction methods (e.g., Wu & Jin, 2021), which often rely on linear assumptions and struggle to correct complex model biases. A key advancement lies in the model's ability to effectively mitigate the issue of "Maritime Continent barrier," ensuring more realistic eastward propagation of the MJO. The framework's robust performance across diverse model outputs underscores its capability as a universal post-processor, which is adaptable to different underlying dynamical products. The design philosophy, which incorporates MJO-specific temporal and spatial scales into the network's kernels, was rigorously validated through ablation studies (Supplementary Figures S1, S2), confirming that this targeted physical guidance is crucial for achieving optimal performance. Most importantly, our explainable AI analysis provides strong evidence that the model's decision-making is spatially congruent with the observed MJO dynamics (correlation > 0.93). This high degree of physical consistency confirms that the network has learned the authentic error characteristics of MJO forecasts rather than merely fitting the training data, thereby providing strong confidence in its generalizability and reliability for future operational applications.

Notwithstanding its demonstrated success, certain limitations of our approach merit consideration and pave the way for future research. First, it is important to recognize that our model, at its core, remains a powerful nonlinear mapper trained on historical data. A key preprocessing step—the removal of low-frequency variability including the long-term trend—effectively filters out the slowly evolving climate change signal. While this isolates the subseasonal MJO signal for learning, it also means our model does not account for the potential modulation of MJO characteristics by global warming, which studies suggest may alter its amplitude, propagation, or frequency (e.g., Rushley et al., 2018). Consequently, the model's ability to reliably correct forecasts during periods dominated by strong, non-stationary external forcing—such as extreme El Niño events or the climatic impact of major volcanic eruptions—may also be limited, as these pose a general challenge for AI models trained on a stationary climate distribution.

Second, our current variable selection (OLR, U850, U200) is grounded in the classical EOF definition of the MJO. However, the flexibility of the AI framework invites exploration beyond this convention. A promising future direction is informed by our improved understanding of MJO physics. Key processes such as the pre-moistening of the lower troposphere, the vertical overturning circulation, and cloud-radiative feedbacks are known to be crucial for MJO initiation and propagation (e.g., Zhang et al., 2020; Jiang et al., 2020). Therefore, incorporating additional atmospheric variables that better represent these processes—such as specific humidity, vertical velocity, and even satellite-derived cloud products—could provide a richer and more physically coherent feature set for the model. This has the potential to unlock further gains in correction skill by allowing the AI to learn from a more complete representation of the MJO's dynamical and thermodynamic environment.

In conclusion, this study successfully bridges the gap between state-of-the-art deep learning and traditional dynamical forecasting for subseasonal prediction. We have developed not merely a skillful AI corrector, but a physically consistent and computationally efficient framework that significantly extends the forecast horizon of the MJO across multiple operational systems. By strategically guiding the model architecture with domain knowledge and rigorously validating its decisions through explainable AI, we have moved beyond the "black box" paradigm towards a more trustworthy and interpretable AI-augmented forecasting paradigm. The pathway outlined here—combining physical priors with data-driven optimization—holds considerable promise for improving predictions of other high-impact climate phenomena, ultimately contributing to more reliable subseasonal-to-seasonal forecasts in a changing climate.

# Data Availability Statement

The atmospheric circulation data used in this study are from the NCEP/DOE Reanalysis-2 (R2) dataset, accessible through the Physical Sciences Laboratory (PSL) at NOAA: https://psl.noaa.gov/data/gridded/data.ncep.reanalysis.pressure.html. The outgoing longwave radiation (OLR) data are obtained from the NOAA Interpolated OLR product, available at: https://psl.noaa.gov/data/gridded/data.olrcdr.interp.html. The subseasonal forecast data were provided by the respective modeling centers (CMA, ECMWF, and NCEP) and can be acquired through their institutional data portals or upon request. We provide the model code and drawing code at: https://github.com/zeaccepted/MJO-Corrector.

# Acknowledgments

We acknowledge the CMA, ECMWF, and NCEP for providing the subseasonal forecast data. The NCEP/DOE Reanalysis-2 and NOAA OLR data were provided by the NOAA PSL. This work was supported by the National Natural Science Foundation of China (Grant No. 42175052, U2442206, 42125503) and the China Meteorological Administration Youth Innovation Team (Grant No. CMA2024QN06).

# Supplementary

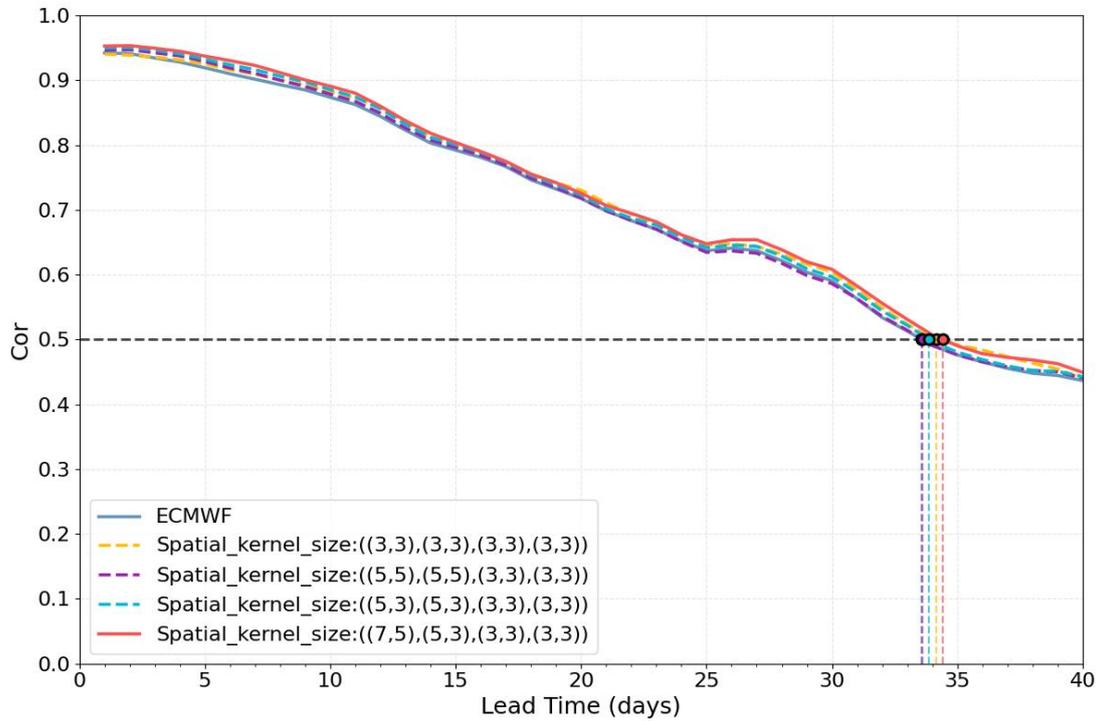

Supplementary Figure S1. Ablation study on 2D convolutional kernel sizes in the U-Net architecture. The bivariate correlation skill is shown for the ECMWF model forecasts corrected by PCC-MJO variants employing different spatial kernel configurations. The kernel sizes in the parentheses (e.g., (7,5), (5,3), (3,3), (3,3)) correspond to the spatial receptive fields of the consecutive layers in the encoder, which are applied in reverse order in the decoder.

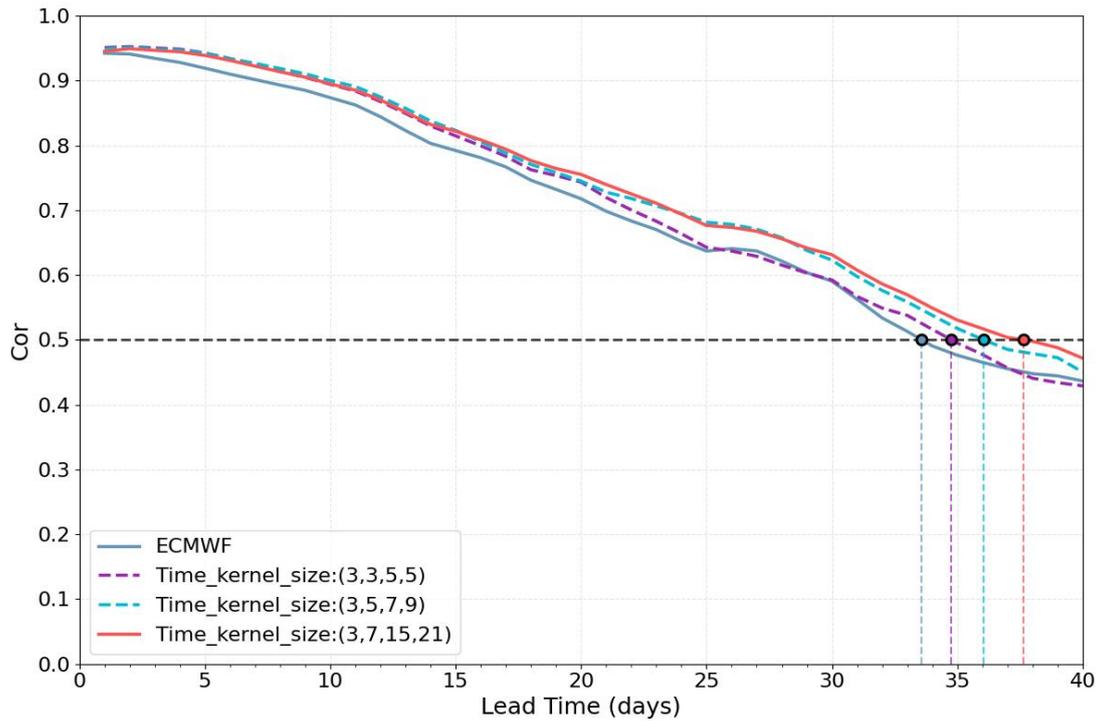

Supplementary Figure S2. Ablation study on temporal convolutional kernel sizes in the 3D U-Net. The bivariate correlation skill is shown for the ECMWF model forecasts corrected by PCC-MJO variants employing different temporal kernel configurations. The kernel sizes in the parentheses (e.g., (3, 7, 15, 21)) correspond to the temporal receptive fields of the consecutive layers in the encoder, which are applied in reverse order in the decoder.

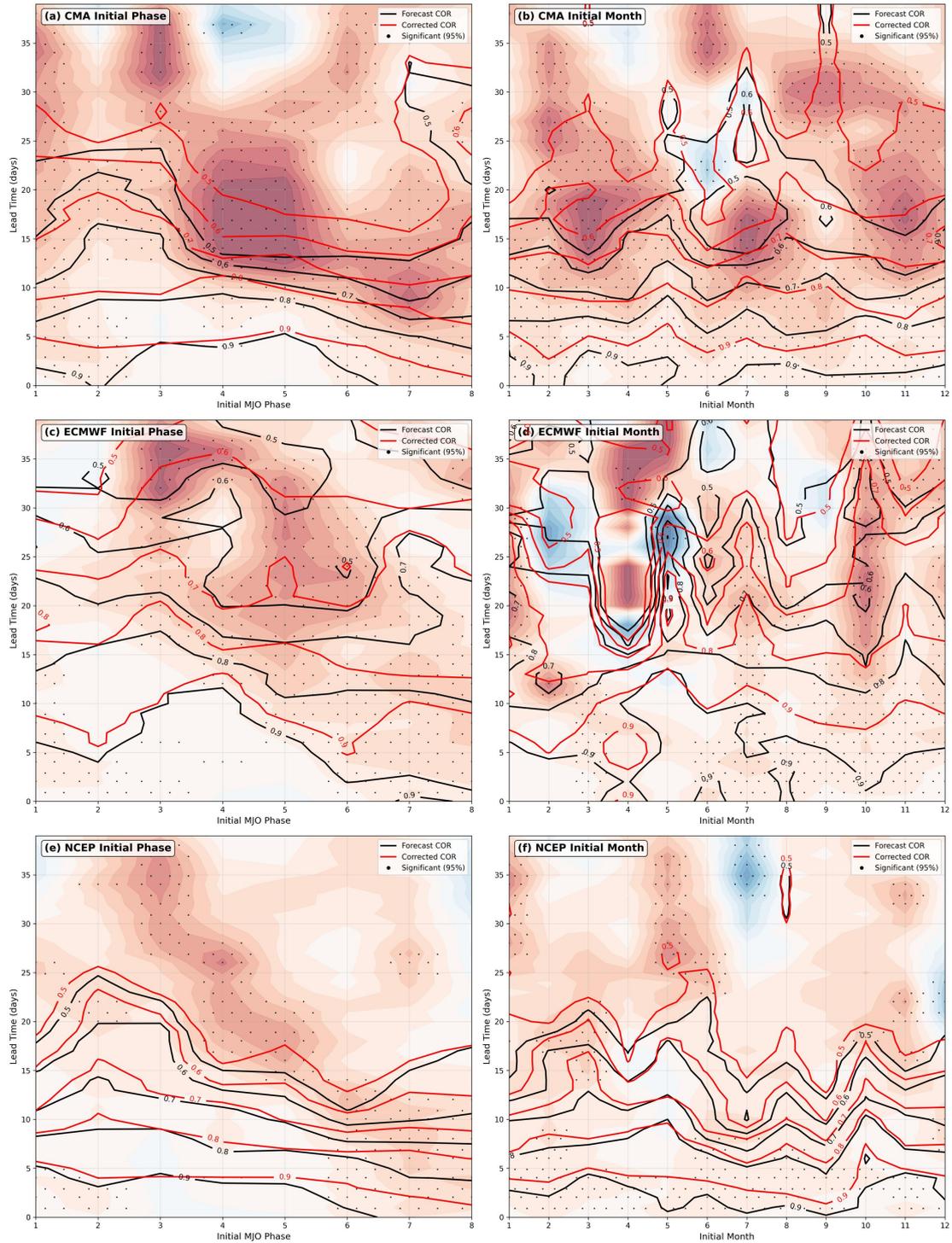

Supplementary Figure S3. Dependence of MJO forecast skill improvement on initial condition. The skill (COR) of raw (black contours) and AI-corrected (red contours) forecasts is shown as a function of initial MJO phase (left column; a, c, e) and initial calendar month (right column; b, d, f), for CMA (a-b), ECMWF (c-d), and NCEP (e-f) models. The shading indicates the difference in COR (corrected minus raw), with red (blue) representing an improvement (degradation). Stippling denotes where the improvement is statistically significant at the 95% confidence level based on Steiger's Z-test. All results are computed

from the test dataset (the last 20% of the data).

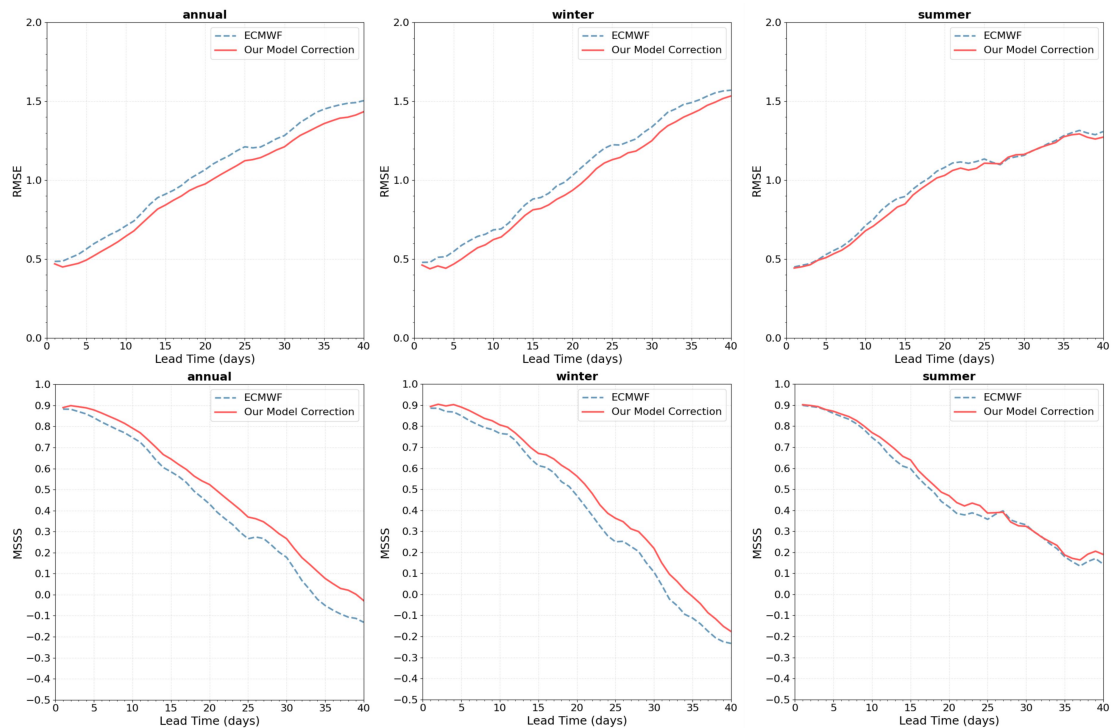

Supplementary Figure S4. Comprehensive evaluation of forecast error and skill improvement using ECMWF data. (a-c) Bivariate root mean square error (RMSE) and (d-f) Murphy's Mean Square Skill Score (MSSS) for raw (blue) and AI-corrected (red) MJO forecasts, evaluated over the (a, d) entire year, (b, e) boreal winter (NDJFM), and (c, f) boreal summer (MJJAS). A lower RMSE and a higher MSSS indicate better forecast performance. The consistent reduction in RMSE and increase in MSSS across all seasons demonstrate the robust skill improvement brought by the AI correction. All results are computed from the test dataset (the last 20% of the data).

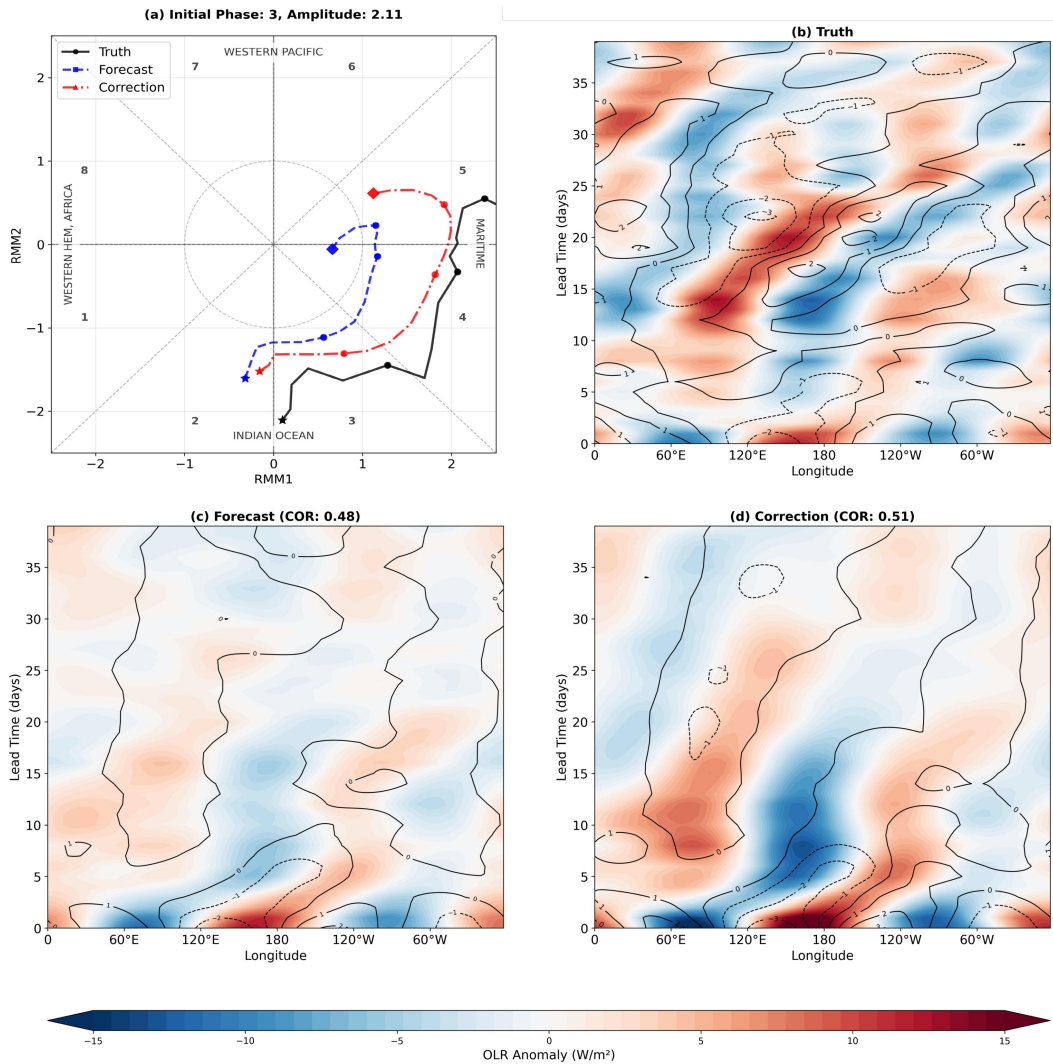

Supplementary Figure S5. Case study of an MJO event initialized in Phase 3 (Amplitude = 2.11), demonstrating forecast improvement after physics-guided design correction. (a) RMM phase-space trajectories from observation (NCEPII, black), raw ECMWF forecast (blue), and AI-corrected forecast (red). (b–d) Hovmöller diagrams of equatorial-averaged (15°S–15°N) OLR anomalies (shading) and 850-hPa zonal wind (U850, contours) for (b) observation, (c) raw forecast, and (d) AI-corrected forecast. The pattern correlation coefficient (PCC) of OLR relative to observations is shown for forecasts.

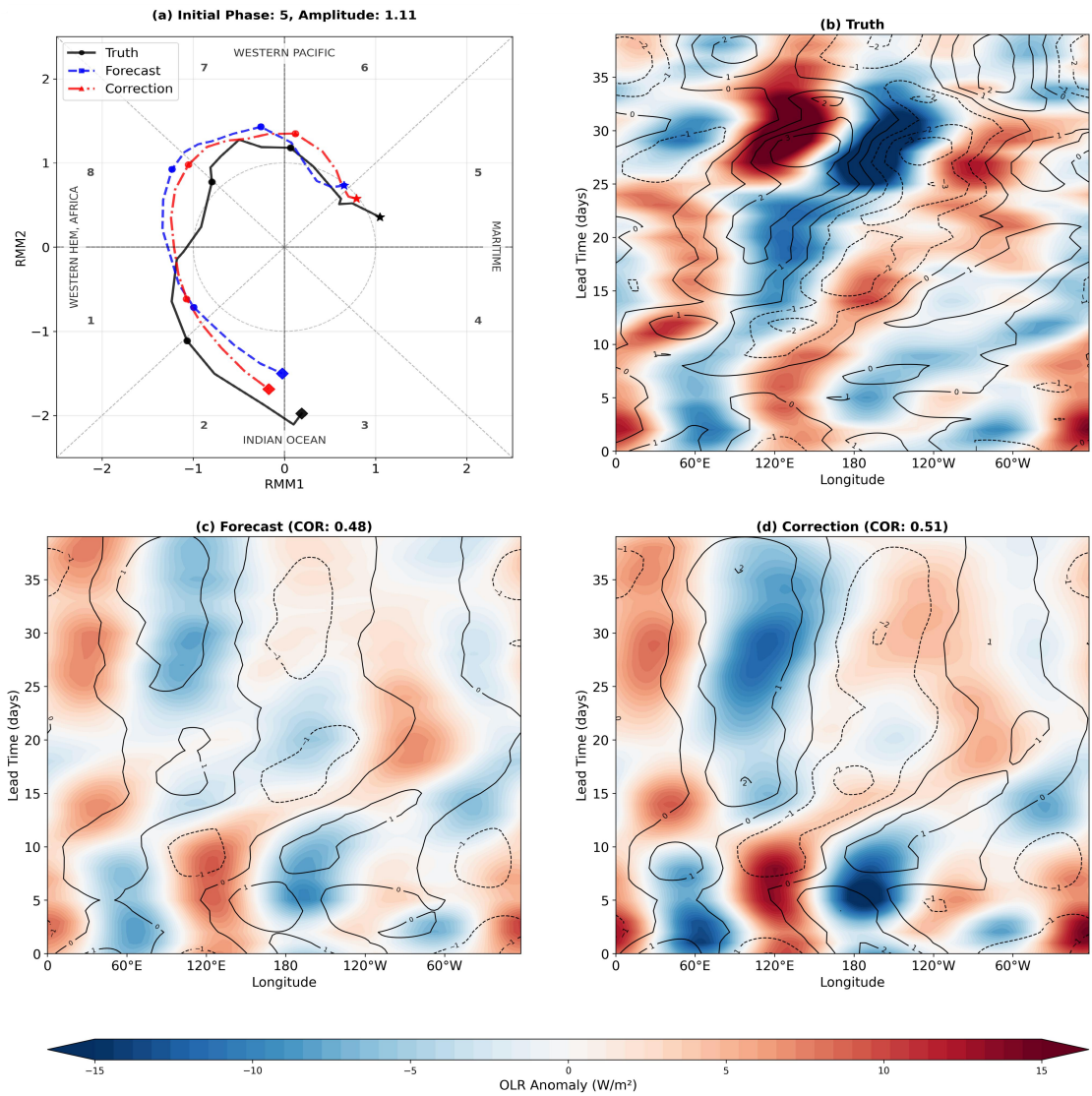

Supplementary Figure S6. As in Supplementary Figure S5, but for an MJO event initialized in Phase 5 (Amplitude = 1.11).

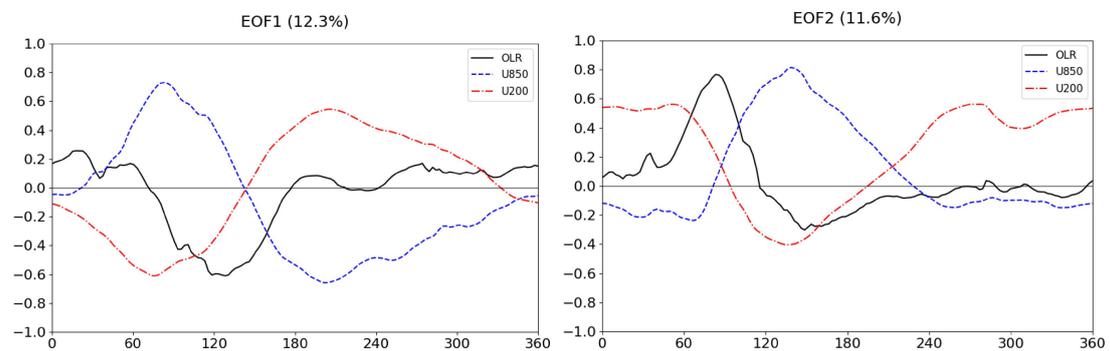

Supplementary Figure S7. Observed spatial patterns of the leading MJO modes. The first two empirical orthogonal function (EOF) modes of the combined fields of outgoing longwave radiation (OLR, black), 850-hPa zonal wind (U850, blue), and 200-hPa zonal wind (U200, red) anomalies over the tropics (15°S–15°N), derived from the ERA5/NCEPII reanalysis. (a)

EOF1 and (b) EOF2 account for 12.3% and 11.6% of the total variance, respectively. These patterns, which form the basis of the RMM index (Wheeler and Hendon, 2004), represent the canonical spatial structures of the MJO and serve as the physical benchmark for interpreting the AI model's behavior.